\documentclass[runningheads,a4paper]{llncs}

\usepackage{amssymb}
\usepackage{graphicx}

\usepackage{url}

\usepackage{amsmath}
\usepackage{booktabs}
\usepackage{color}
\usepackage{hyperref}
\usepackage{multirow}
\usepackage{amsmath}
\usepackage{cite}

\DeclareMathOperator*{\argmax}{\arg\!\max}

\begin{document}

\mainmatter  

\title{Communicative Reinforcement Learning Agents for Landmark Detection in Brain Images
}
\titlerunning{Communicative Reinforcement Learning Agents}


\author{
Guy Leroy%
\and Daniel Rueckert
\and Amir Alansary 
}


\authorrunning{
G. Leroy et al.
}


\institute{
Imperial College London, London, UK
}

\toctitle{Lecture Notes in Computer Science}
\tocauthor{Authors' Instructions}
\maketitle


\begin{abstract}
Accurate detection of anatomical landmarks is an essential step in several medical imaging tasks.
We propose a novel communicative multi-agent reinforcement learning (C-MARL) system to automatically detect landmarks in 3D medical scans. 
C-MARL enables the agents to learn explicit communication channels, as well as implicit communication signals by sharing certain weights of the architecture among all the agents.
The proposed approach is evaluated on two brain imaging datasets from adult magnetic resonance imaging (MRI) and fetal ultrasound scans. 
Our experiments show that involving multiple cooperating agents by learning their communication with each other outperforms previous approaches using single agents.
\end{abstract}


\section{Introduction}\label{sec_introduction}
Robust and fast landmark localization is an essential step in medical imaging analysis applications including biometric measurements of anatomical structures \cite{payer2016regressing}, registration of 3D volumes \cite{lian2018hierarchical} and extraction of 2D clinical standard planes \cite{li2018fast}. 
Manual labeling of such landmarks is often a time-consuming and tedious task, which is also error-prone and requires human experts.
Developing accurate and automatic detection methods will help reduce the human error and speed the diagnosis process.
Recent advances in reinforcement learning (RL) have shown a significant contribution to clinical applications such as automated medical diagnosis, object localization, and landmark detection \cite{yu2019reinforcement}.
RL enables learning from reward signals that guide the agent towards the target solution in sequential steps during training. 
It learns to perform a non-exhaustive search without using the full 3D image as an input. 
RL can be data efficient by using the same 3D image for training with different starting points and states. 
RL has proven to achieve the best performance for landmark detection outperforming supervised methods \cite{ghesu2016artificial, ghesu2017multi, alansary2019evaluating}.

\textbf{Related Work:}
Previous works detecting anatomical landmarks have examined approaches including statistical shape priors, regression forests \cite{oktay2016stratified}, \cite{gauriau2015multi}, Hough voting \cite{basher2019hippocampus}, supervised convolutional neural network (CNN) \cite{li2018fast} and attention-based autoencoder \cite{zhong2019attention}.
With the recent advances of deep RL, Ghesu et al. \cite{ghesu2016artificial} introduced the application of RL to detect anatomical landmarks by learning sequential actions towards the target landmark, while outperforming supervised methods.
Alansary et al. \cite{alansary2019evaluating} then evaluated multiple deep Q-network (DQN) variants for the detection task, namely DQN \cite{mnih2015human}, double DQN \cite{van2016deep}, dueling DQN \cite{wang2016dueling}, and double dueling DQN. 
They also incorporated hierarchical steps with the multi-scale search strategy, which significantly decreased the search time.
Multi-scale agents have proven to outperform fixed-scale agents for detecting the majority of landmarks \cite{ghesu2017multi, alansary2019evaluating}. 
Vlontzos et. al \cite{vlontzos2019multiple} proposed the first multi-agent system for landmark detection, where the agents communicate efficiently by sharing the convolutional weights of the CNN model.
Furthermore, RL has been utilized in various medical applications such as the detection of standardized view planes in MRI scans \cite{alansary2018automatic}, organ localization in CT scans \cite{navarro2020deep}, and re-identifying the location of brain landmarks in pre- and post-operative images \cite{waldmannstetter2020reinforced}.

\textbf{Contributions:}
(I) We propose a novel communicative multi-agent reinforcement learning for multiple landmarks detection. (II) Experiments are evaluated on two different brain imaging datasets from adult MRI and fetal ultrasound, outperforming previously published RL state-of-the-art results. (III) The implementation of the code is publicly available.

\section{Background}
\label{sec_background}
Reinforcement learning (RL) is a sub-field of machine learning (ML), which lies under the bigger umbrella of artificial intelligence (AI). 
Inspired from behavioral psychology and neuroscience \cite{sutton2018reinforcement}, an RL agent takes actions within an environment and receives updated states with associated rewards during training. 
These reward signals guide the agent to take correct actions towards the target solution, and penalize otherwise. 
Thus, the agent learns a policy $\pi$ directly from high-dimensional inputs.
In most modern applications, including ours, agents will not have total knowledge of all environment's states. 
This is referred to as a partially observable Markov decision process (MDP).
RL offers an efficient solution to deal with the MDP by learning a policy that maximizes the total rewards. For instance, Q-learning \cite{watkins1992q} seeks to find a q-value that measures the quality of taking an action $a$ given a current state $s$ by learning a policy $\pi$ that maximizes the total reward during training.
Mnih et al. \cite{mnih2015human} proposed to approximate these q-values using a deep neural network ($\theta$), named DQN.
The $Q$-function is based on the Bellman equation \cite{bellman1966dynamic}, and defined as the expected discounted cumulative rewards:
\begin{equation}
    Q^\pi(s_t,a_t)=E_\pi[\sum_{k=0}^\infty \gamma^k r_{t+k+1}|s_t,a_t],
    \label{eqn_qfunction}
\end{equation}
where $s_t$ and $a_t$ represent the state and action at step $t$. $\gamma^k$ is the discount factor at $k$-th future state.
DQN introduces another target network $\hat{Q}$ that stabilizes the training, and reduce the overestimation of the maximum Q-value \cite{mnih2015human}. 
Whereas at every predefined interval during training, the weights $\theta$ of the Q-network are copied to the target network $\hat{\theta}$. 
The DQN loss function is defined as:
\begin{equation}
    L_i(\theta_i)=E_{s,a,r,s'}\left[ \left( r+\gamma \max_{a'}\hat{Q}(s',a';\hat{\theta}_i)-Q(s,a;\theta_i) \right)^2\right],
    \label{eqn_DQN}
\end{equation}
where $s'$ and $a'$ are the next state and action.
Van Hasselt et al. \cite{van2016deep} introduced a modification to the DQN loss function to decouple the selected action from the target network, known as double DQN. This changes the loss function to,
\begin{equation}
    L_i(\theta_i)=E_{s,a,r,s'}\left[ \left( r+\gamma \hat{Q}(s',\argmax_{a'}Q(s',a';\theta);\hat{\theta}_i)-Q(s,a;\theta_i) \right)^2\right].
    \label{eqn_doubleDQN}
\end{equation}
The dueling network \cite{wang2016dueling} uses the hypothesis that Q-values are only important in key states. It has two sequences of fully connected layers to separately estimate state-values and the advantages for each action as scalars. 

Alansary et al. \cite{alansary2019evaluating} have shown that the optimal DQN architecture depends on each landmark, where there was no overall best architecture for all landmarks. 
Thus, we use the double DQN as a baseline architecture.
\section{Methods}
\label{methods}
\begin{figure}[!htb]
	\centering
	\includegraphics[width=0.5\linewidth]{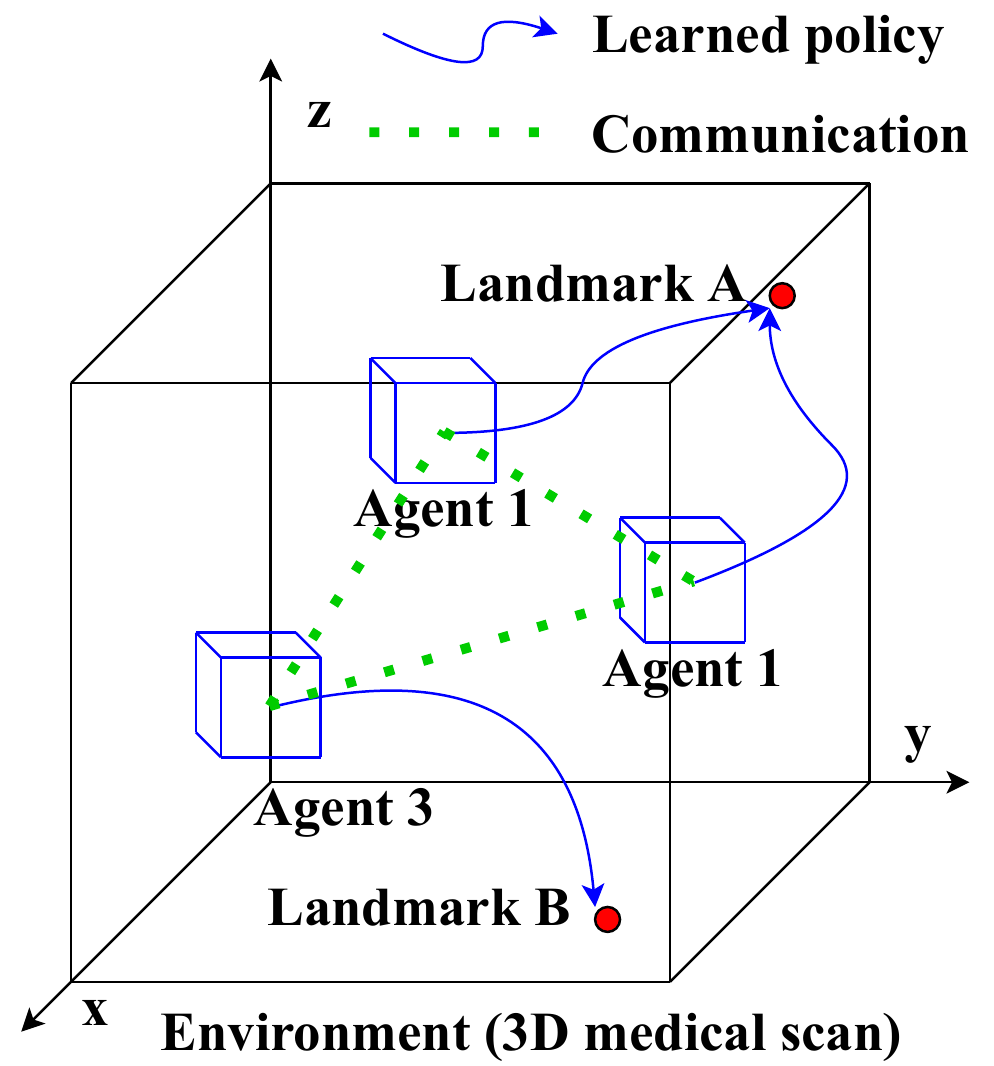}
	\caption{A schematic diagram of the proposed multi-agents interacting with the 3D image environment $E$. At each step, each agent takes an action towards a target landmark. The learned policy is formed by the path between the starting points and the target landmarks after taking the sequential actions.}
	\centering
	\label{fig_environment}
\end{figure}

In this work, we propose a communicative DQN-based RL agents for the detection of anatomical landmarks in brain images.
These agents are designed to learn by communication during their search for different landmarks in 3D medical scans. 
This is motivated by the fact that anatomical landmarks are usually spatially correlated in the brain. 
Figure \ref{fig_environment} demonstrates a schematic visualization of these navigating agents in a 3D scan or environment $E$.

\textbf{States:}
Each state $s$ is defined as a region of interest (ROI) of size $45\times45\times45$ voxels, and centered around each agent. To improve the network's stability and convergence, it takes as an input a history of the last $4$ states \cite{mnih2015human}. Each agent starts at a random location within the $80\%$ of the inner region of the image at the beginning of each episode. An agent terminates navigating when it finds the target landmark. During inference the terminal state is triggered when the agent oscillates around a target point. 

\textbf{Action Space:}
It is defined based on the six directions in the 3D Cartesian coordinates, namely left, right, up, down, forward or backward. Similar to \cite{alansary2019evaluating}, we adopt a multi-scale search strategy with hierarchical steps by reducing the step and ROI size when the agent oscillates around a target point. 
We use three levels of scales $\{3,2,1\}$ mm.
The episode is terminated when all agents reach their terminal states at the 1mm scale.

\textbf{Rewards:}
First, we calculate the Euclidean distance between the current point of interest and target landmark $d_t$, and between the point of interest of the previous step and the target landmark $d_{t-1}$. 
The reward signal is then calculated using the difference between $d_{t-1}$ and $d{t}$, and clipped between -1 and 1. 
This ensures that positive rewards are given, if the movements of the agent are towards the target solution. 

\begin{figure}[!hbt]
	\centering
	\includegraphics[width=\linewidth]{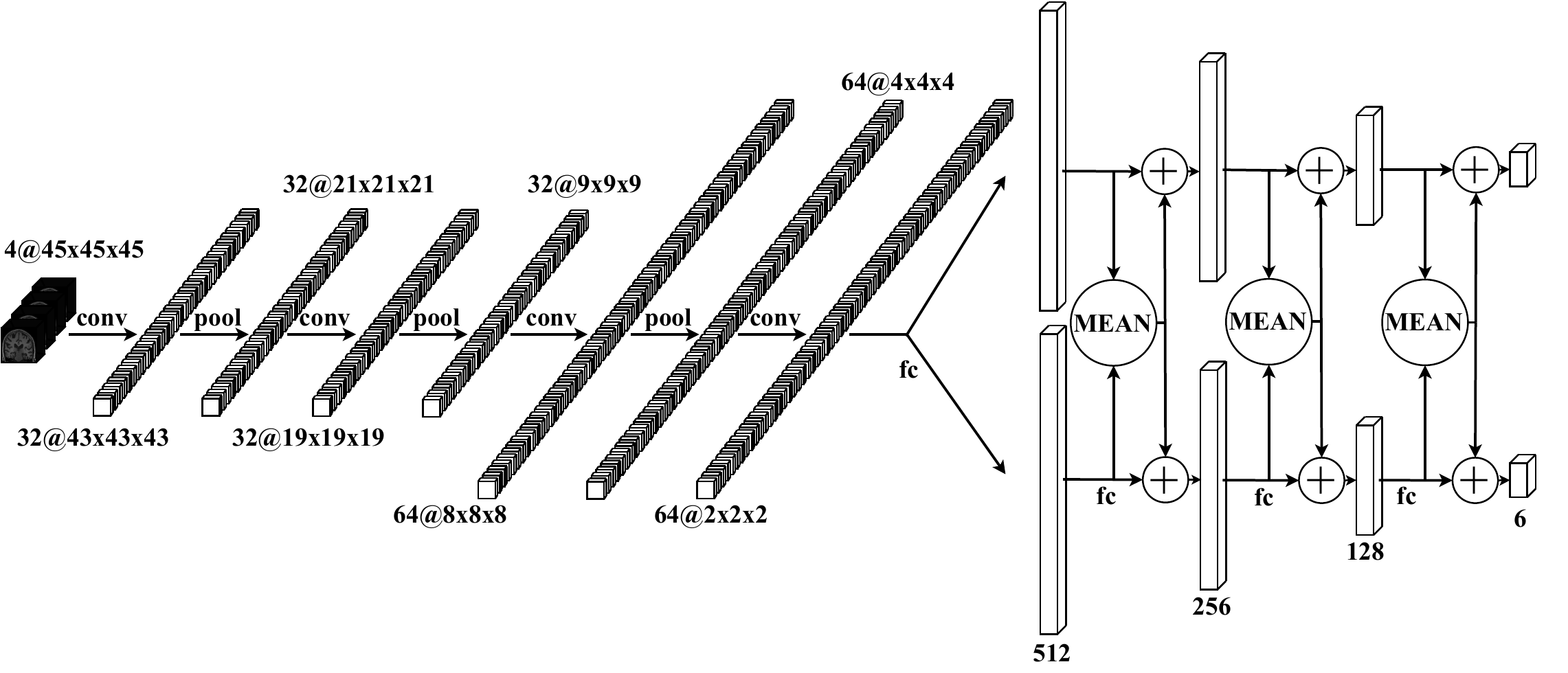}
	\caption{The proposed C-MARL architecture for anatomical landmark detection. Here is an example of two agents sharing the same convolutional layers. They learn to communicate by averaging the output of the FC layer of each agent, which is then concatenated to the input of the next FC layer.}
	\label{fig_commnet}
\end{figure}

\textbf{Communicative Agents:}
We leverage two types of communications between the agents. Implicit communication is learned by sharing the convolutional layers of the model among all the agents \cite{vlontzos2019multiple}. Besides, communication signals are learned explicitly by sharing communication channels in the fully connected (FC) layers \cite{sukhbaatar2016learning}. This is implemented by averaging the output of each FC layer for each agent, which is then concatenated with the input of the next FC layer, as seen in Fig. \ref{fig_commnet}. 

\textbf{Network Architecture:}
Figure \ref{fig_commnet} shows the architecture of the proposed C-MARL model, which takes as an input a tensor of size $\texttt{number\_agents}\times4\times45\times45\times45$. 
It consists of four 3D convolutional and three 3D max pooling layers, followed by four FC layers.
Whereas the convolutional layers are shared between all the agents, and each agent has its own FC layer.
The output of all FC layers of each agent are averaged and concatenated with the input of the next FC layer.
The size of the last FC layer is the same size of the action space.
Finally, the model is trained using Eqn. \ref{eqn_doubleDQN}.

\section{Experiments}
\begin{figure}[!hbt]
	\centering
	\includegraphics[width=\linewidth]{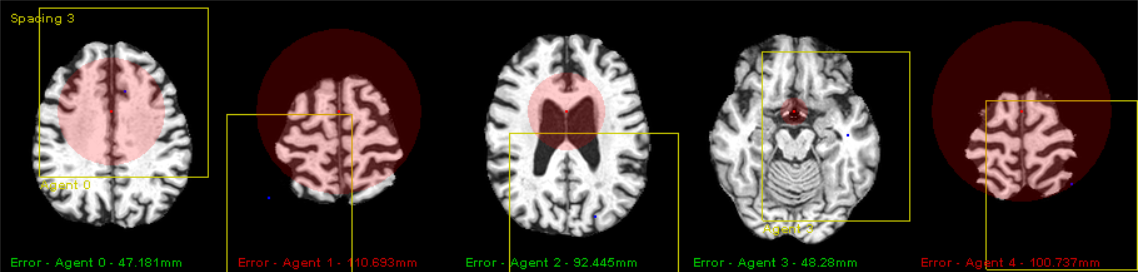}
	\caption{An example of our proposed C-MARL system consisting of 5 agents. These agents are looking for 5 different landmarks in a brain MRI scan. 
	Each agent's ROI is represented by a yellow box and centered around a blue point, while the red point is the target landmark. 
	ROI is sampled with $3$ mm spacing at the beginning of every episode.
	The length of the circumference of red disks denotes the distance between the current and target landmarks in $z$-axis. 
	}
	\centering
	\label{fig_5agents}
\end{figure}

The performance of the proposed C-MARL agents for anatomical landmark detection is tested on two brain imaging datasets, and evaluated against a single RL agent \cite{alansary2019evaluating} and multi-agents that share only their convolutional layers (Collab-DQN) \cite{vlontzos2019multiple}.
Clinical experts manually annotated all selected landmarks using three orthogonal views.
We have randomly split both datasets into train ($70\%$), validation ($15\%$) and test ($15\%$) subsets. Best model is selected during training based on the best accuracy on the validation subset. 
The Euclidean distance error between the detected and target landmarks is used to measure the reported accuracy. 
The agents follow an $\epsilon$-greedy policy, where each agent can take a random action step uniformly sampled from the action space with an initial probability of $\epsilon=1$ to $\epsilon=0.1$, instead of selecting the step with the highest Q-value. 
During testing, agents follow a full greedy policy with $\epsilon=0$. 
The episode ends when all agents oscillate at the smallest scale, or after a predefined maximum number of $200$ steps. 
Figure \ref{fig_5agents} shows C-MARL performing with five agents to detect five different landmarks from a brain MRI scan. 

\subsection{Results}
\textbf{Experiment (I):}
We use $832$ T1-weighted $1.5$T MRI brain scans from the Alzheimer’s disease neuroimaging initiative (ADNI)\footnote{http://adni.loni.usc.edu}. 
All brain images are skull-stripped, and have an isotropic  $1$ $mm^3$ voxel size.
The selected subjects include patients with cognitively normal (CN), mild cognitive impairment (MCI), and early Alzheimer’s disease (AD).
We select $8$ landmarks, namely the anterior commissure (AC), the posterior commissure (PC), the outer aspect, the inferior tip and inner aspect of the splenium of the corpus callosum (SCC), the outer and inner aspect of the Genu of corpus callosum (GCC), and the superior aspect of pons.

Table \ref{results_test_brain_table} demonstrates the performance of the different approaches, whereas C-MARL with three agents achieves the best accuracy for all the three selected landmarks. 
The table also shows experiments using larger number of agents (five and eight).
These experiments result in a decrease in the accuracy in most of the landmarks compared to the results using three agents.
Thus, intuitively, increasing the number of agents may require architectures with a bigger capacity to be able to learn more communications.
Another explanation can be that adding more landmarks, that are not strongly correlated, may affect the detection accuracy.  
\begin{table}[!hbt]
\begin{center}
\resizebox{\columnwidth}{!}{%
\begin{tabular}{l|c|c|c|c|c|c|c|}
\cline{2-8}
 & \multirow{2}{*}{\textbf{\begin{tabular}[c]{@{}l@{}}Single \\ agent \cite{alansary2019evaluating}\end{tabular}}} & \multicolumn{3}{c|}{\textbf{Collab-DQN \cite{vlontzos2019multiple}}} & \multicolumn{3}{c|}{\textbf{C-MARL}} \\ \cline{1-1} \cline{3-8} 
\multicolumn{1}{|l|}{\textbf{Landmark}} &  & \textbf{3 agents} & \textbf{5 agents} & \textbf{8 agents} & \textbf{3 agents} & \textbf{5 agents} & \textbf{8 agents} \\ \hline
\multicolumn{1}{|l|}{\textbf{AC}} & 1.14$\pm$0.53 & 1.16$\pm$0.59 & 1.13$\pm$0.64 & 1.21$\pm$0.92 & \textbf{1.04$\pm$0.58} & 1.12$\pm$0.65 & 1.84$\pm$0.91 \\ \hline
\multicolumn{1}{|l|}{\textbf{PC}} & 1.18$\pm$0.55 & 1.25$\pm$0.57 & 1.19$\pm$0.61 & 1.22$\pm$0.93 & \textbf{1.13$\pm$0.66} & 1.25$\pm$0.55 & 1.38$\pm$0.64 \\ \hline
\multicolumn{1}{|l|}{\textbf{Outer SCC}} & 1.47$\pm$0.64 & 1.38$\pm$0.75 & 1.51$\pm$0.77 & 1.46$\pm$0.90 & \textbf{1.35$\pm$0.66} & 1.62$\pm$0.79 & 5.20$\pm$13.49 \\ \hline
\multicolumn{1}{|l|}{\textbf{Inferior SCC}} & 2.40$\pm$1.13 & - & \textbf{1.39$\pm$0.85} & 1.53$\pm$0.87 & - & 1.50$\pm$0.89 & 1.87$\pm$1.28 \\ \hline
\multicolumn{1}{|l|}{\textbf{Inner SCC}} & 1.46$\pm$0.73 & - & 1.53$\pm$0.97 & 2.09$\pm$3.65 & - & \textbf{1.53$\pm$0.76} & 3.56$\pm$9.42 \\ \hline
\end{tabular}
}
\caption{Comparison between single, multiple, and communicative agents for landmark detection in brain MRIs. Distance errors are in $mm$.}
\label{results_test_brain_table}
\end{center}
\end{table}

\textbf{Experiment (II):}
We use $72$ subjects of 3D fetal head ultrasound scans from the iFIND project\footnote{http://www.ifindproject.com}.
All images are resampled to isotropic voxel size with average dimensions of $324\times207\times279$ voxels.
We select the right and left cerebellum (RC and LC respectively), the cavum septum pellucidum (CSP) and the center and anterior head (CH and AH respectively) landmarks.

Table \ref{results_test_fetal_table} shows multiple agents have a lower distance error across all fetal landmarks, while C-MARL significantly outperforms the other methods for detecting the CSP and CH. 
Similar to the previous experiment, increasing the number of agents did not necessarily improve the detection accuracy. 
However, the AH landmark has significantly benefited from increasing the number of agents.
In this experiment, results show that multi-agent system is superior in all landmarks, but rather suggest the best architecture depends on the landmark.
\begin{table}[!htb]
\begin{center}
\resizebox{\columnwidth}{!}{%
\begin{tabular}{l|c|c|c|c|c|c|c|}
\cline{2-8}
 & \multirow{2}{*}{\textbf{\begin{tabular}[c]{@{}l@{}}Single \\ agent \cite{alansary2019evaluating}\end{tabular}}} & \multicolumn{3}{c|}{\textbf{Collab-DQN \cite{vlontzos2019multiple}}} & \multicolumn{3}{c|}{\textbf{C-MARL}} \\ \cline{1-1} \cline{3-8} 
\multicolumn{1}{|l|}{\textbf{Landmark}} &  & \textbf{3 agents} & \textbf{5 agents} & \textbf{8 agents} & \textbf{3 agents} & \textbf{5 agents} & \textbf{8 agents} \\ \hline
\multicolumn{1}{|l|}{\textbf{RC}} & 7.23$\pm$3.54 & \textbf{2.73$\pm$1.71} & 4.20$\pm$3.76 & 3.39$\pm$2.36 & 6.53$\pm$4.21 & 4.06$\pm$2.95 & 4.75$\pm$3.28 \\ \hline
\multicolumn{1}{|l|}{\textbf{LC}} & 4.37$\pm$1.45 & \textbf{4.20$\pm$2.87} & 5.98$\pm$8.58 & 5.42$\pm$4.50 & 5.10$\pm$3.66 & 4.43$\pm$32.26 & 4.64$\pm$3.16 \\ \hline
\multicolumn{1}{|l|}{\textbf{CSP}} & 9.90$\pm$3.13 & 5.18$\pm$2.05 & 8.02$\pm$5.34 & 5.74$\pm$5.07 & 5.78$\pm$3.04 & \textbf{5.13$\pm$3.51} & 7.08$\pm$4.13 \\ \hline
\multicolumn{1}{|l|}{\textbf{CH}} & 29.43$\pm$17.83 & - & 14.45$\pm$5.25 & 16.83$\pm$12.54 & - & \textbf{13.00$\pm$4.97} & 16.29$\pm$8.94 \\ \hline
\multicolumn{1}{|l|}{\textbf{AH}} & 5.73$\pm$2.88 & - & 8.11$\pm$5.22 & \textbf{4.10$\pm$2.26} & - & 4.33$\pm$2.96 & 8.89$\pm$4.91 \\ \hline
\end{tabular}
}
\caption{Comparison between single, multiple, and communicative agents for landmark detection in fetal head ultrasound. Distance errors are in $mm$.}
\label{results_test_fetal_table}
\end{center}
\end{table}

\textbf{Experiment (III):}
The previous experiments are conducted in the scenario of using a single agent for the detection of one landmark. 
In this experiment, we proceed to evaluate the performance of using multi-agents for detecting the same single landmark. 
The final location of the agents are averaged at the end of an episode. 
To give a baseline, we include a column for five single agents looking for the same landmark in parallel. 
We report the results on a selected landmark from each dataset used in the previous two experiments, namely AC and CSP. 
Table \ref{results_all_on_one_table} shows C-MARL's results are much better than in any of the previous methods. 
Parallel single agents are not significantly better than the results with only one agent. 

\begin{table}[!htb]
\begin{center}
\begin{tabular}{|l|c|c|c|}
\hline
\textbf{Landmarks} & \textbf{\begin{tabular}[c]{@{}l@{}}Single agents \cite{alansary2019evaluating}\end{tabular}} & \textbf{Collab-DQN \cite{vlontzos2019multiple}} & \textbf{C-MARL} \\ \hline
\textbf{AC} & 0.97$\pm$0.40 & 0.81$\pm$0.36 & \textbf{0.75$\pm$0.34} \\ \hline
\textbf{CSP} & 10.43$\pm$4.28 & 6.66$\pm$4.19 & \textbf{5.10$\pm$4.25} \\ \hline
\end{tabular}
\caption{Results from using five agents looking for the same landmark. Distance error are in $mm$.}
\label{results_all_on_one_table}
\end{center}
\end{table}

\textbf{Experiment (IV):}
We further evaluate using multi agents for detecting multiple landmarks, where each single landmark have multiple agents.
In this experiment, we train four agents to detect the AC and PC landmarks, where each landmark has two dedicated agents.
Similar to the previous experiment, to give a baseline, we compare with four non communicating agents as a baseline.
Table \ref{results_hybrid_table} shows that C-MARL agents perform better than the baseline, but worse than using five agents for a single landmark from Experiment (III).
Finally, these experiments show that multiple cooperative agents trained to detect one single landmark can outperform the same number of agents detecting different landmarks.
\begin{table}[!htb]
\begin{center}
\begin{tabular}{|l|c|c|}
\hline
\textbf{Landmarks} & \textbf{\begin{tabular}[c]{@{}l@{}}Single agents \cite{alansary2019evaluating}\end{tabular}} & \textbf{C-MARL} \\ \hline
\textbf{AC} & 1.17$\pm$0.61 & \textbf{0.95$\pm$0.43} \\ \hline
\textbf{PC} & 1.12$\pm$0.55 & \textbf{0.97$\pm$0.46} \\ \hline
\end{tabular}
\caption{Results from using two pairs of agent looking for two landmarks (four agents in total). Distance error are in $mm$.}
\label{results_hybrid_table}
\end{center}
\end{table}

\textbf{Implementation:}
We run each experiment for four days, but each would converge usually after one or two days. 
We used Nvidia Tesla or Nvidia GeForce GTX Titan Xp with 12GB RAM, using CUDA v10.0.130 and Torch v1.4. 
A 24-core/48 thread Intel Xeon CPU was used with 256GB RAM. 
In four days, collab-DQN ran 30k episodes while our proposed method only ran 20k episodes. 
The memory space during training is mostly driven up by the memory buffer, which we set to $\frac{100,000}{\#agents}$ episodes.
As for the model's size, more agents take up more space and communication channels are added on the collab-DQN's architecture. 
More precisely, our model size is $5,504,759$ and $8,144,365$ bytes for three and five agent respectively, while for collab-DQN it is $3,529,451$ and $4,852,185$ bytes. 
For comparison, three single agents working independently have model size $2,206,723\times3=6,620,169$ bytes and for five single agents it is $2,206,723\times5=11,033,615$ bytes. 
This shows multi-agent models greatly reduce the models' trainable parameters. 
For the testing speed, our method takes around 2.5 and 4.9 seconds per episode for three and five agents respectively and those figures are 2.2 and 4.2 seconds for collab-DQN. 
The code is publicly available on Github, \url{https://github.com/gml16/rl-medical}.

\section{Conclusion} 
We introduced a communicative multi-agent reinforcement learning (C-MARL) system for detecting multiple anatomical landmarks from brain medical images. 
Multi-agents share the weights of the convolutional layers to learn implicit communications.
They also learn explicit communication channels calculated from the output of their fully connect layers, which are then shared among them by concatenating to the input of the following fully connected layers. 
C-MARL was evaluated on adult brain MRI and fetal head ultrasound, outperforming single- and multi-agents approaches.

\textbf{Future Work:}
The optimal number of agents and combination of landmarks will be further investigated.
It will be also interesting to research weighted communication channels based on nearby agents to reduce noise from distant landmarks.
We will incorporate more complex communication channels, e.g. skip connections and temporal units. 
Another direction is to investigate competitive approaches for communication instead of collaboration between the agents.


\bibliographystyle{splncs04}
\bibliography{references}

\begin{thebibliography}{10}
\providecommand{\url}[1]{\texttt{#1}}
\providecommand{\urlprefix}{URL }
\providecommand{\doi}[1]{https://doi.org/#1}

\bibitem{alansary2018automatic}
Alansary, A., Le~Folgoc, L., Vaillant, G., Oktay, O., Li, Y., Bai, W.,
  Passerat-Palmbach, J., Guerrero, R., Kamnitsas, K., Hou, B., et~al.:
  Automatic view planning with multi-scale deep reinforcement learning agents.
  In: International Conference on Medical Image Computing and Computer-Assisted
  Intervention. pp. 277--285. Springer (2018)

\bibitem{alansary2019evaluating}
Alansary, A., Oktay, O., Li, Y., Le~Folgoc, L., Hou, B., Vaillant, G.,
  Kamnitsas, K., Vlontzos, A., Glocker, B., Kainz, B., et~al.: Evaluating
  reinforcement learning agents for anatomical landmark detection. Medical
  image analysis  \textbf{53},  156--164 (2019)

\bibitem{basher2019hippocampus}
Basher, A., Choi, K.Y., Lee, J.J., Lee, B., Kim, B.C., Lee, K.H., Jung, H.Y.:
  Hippocampus localization using a two-stage ensemble {Hough} convolutional
  neural network. IEEE Access  \textbf{7},  73436--73447 (2019)

\bibitem{bellman1966dynamic}
Bellman, R.: Dynamic programming. Science  \textbf{153}(3731),  34--37 (1966)

\bibitem{gauriau2015multi}
Gauriau, R., Cuingnet, R., Lesage, D., Bloch, I.: Multi-organ localization with
  cascaded global-to-local regression and shape prior. Medical image analysis
  \textbf{23}(1),  70--83 (2015)

\bibitem{ghesu2016artificial}
Ghesu, F.C., Georgescu, B., Mansi, T., Neumann, D., Hornegger, J., Comaniciu,
  D.: An artificial agent for anatomical landmark detection in medical images.
  In: International conference on medical image computing and computer-assisted
  intervention. pp. 229--237. Springer (2016)

\bibitem{ghesu2017multi}
Ghesu, F.C., Georgescu, B., Zheng, Y., Grbic, S., Maier, A., Hornegger, J.,
  Comaniciu, D.: Multi-scale deep reinforcement learning for real-time
  3d-landmark detection in ct scans. IEEE transactions on pattern analysis and
  machine intelligence  \textbf{41}(1),  176--189 (2017)

\bibitem{li2018fast}
Li, Y., Alansary, A., Cerrolaza, J.J., Khanal, B., Sinclair, M., Matthew, J.,
  Gupta, C., Knight, C., Kainz, B., Rueckert, D.: Fast multiple landmark
  localisation using a patch-based iterative network. In: International
  Conference on Medical Image Computing and Computer-Assisted Intervention. pp.
  563--571. Springer (2018)

\bibitem{lian2018hierarchical}
Lian, C., Liu, M., Zhang, J., Shen, D.: {Hierarchical fully convolutional
  network for joint atrophy localization and Alzheimer's Disease diagnosis
  using structural MRI}. IEEE transactions on pattern analysis and machine
  intelligence  (2018)

\bibitem{mnih2015human}
Mnih, V., Kavukcuoglu, K., Silver, D., Rusu, A.A., Veness, J., Bellemare, M.G.,
  Graves, A., Riedmiller, M., Fidjeland, A.K., Ostrovski, G., et~al.:
  Human-level control through deep reinforcement learning. nature
  \textbf{518}(7540),  529--533 (2015)

\bibitem{navarro2020deep}
Navarro, F., Sekuboyina, A., Waldmannstetter, D., Peeken, J.C., Combs, S.E.,
  Menze, B.H.: {Deep Reinforcement Learning for Organ Localization in CT}.
  arXiv preprint arXiv:2005.04974  (2020)

\bibitem{oktay2016stratified}
Oktay, O., Bai, W., Guerrero, R., Rajchl, M., de~Marvao, A., O’Regan, D.P.,
  Cook, S.A., Heinrich, M.P., Glocker, B., Rueckert, D.: Stratified decision
  forests for accurate anatomical landmark localization in cardiac images. IEEE
  transactions on medical imaging  \textbf{36}(1),  332--342 (2016)

\bibitem{payer2016regressing}
Payer, C., {\v{S}}tern, D., Bischof, H., Urschler, M.: {Regressing heatmaps for
  multiple landmark localization using CNNs}. In: International Conference on
  Medical Image Computing and Computer-Assisted Intervention. pp. 230--238.
  Springer (2016)

\bibitem{sukhbaatar2016learning}
Sukhbaatar, S., Fergus, R., et~al.: Learning multiagent communication with
  backpropagation. In: Advances in neural information processing systems. pp.
  2244--2252 (2016)

\bibitem{sutton2018reinforcement}
Sutton, R.S., Barto, A.G.: {Reinforcement learning: An introduction} (2018)

\bibitem{van2016deep}
Van~Hasselt, H., Guez, A., Silver, D.: Deep reinforcement learning with double
  q-learning. In: Thirtieth AAAI conference on artificial intelligence (2016)

\bibitem{vlontzos2019multiple}
Vlontzos, A., Alansary, A., Kamnitsas, K., Rueckert, D., Kainz, B.: Multiple
  landmark detection using multi-agent reinforcement learning. In:
  International Conference on Medical Image Computing and Computer-Assisted
  Intervention. pp. 262--270. Springer (2019)

\bibitem{waldmannstetter2020reinforced}
Waldmannstetter, D., Navarro, F., Wiestler, B., Kirschke, J.S., Sekuboyina, A.,
  Molero, E., Menze, B.H.: Reinforced redetection of landmark in pre-and
  post-operative brain scan using anatomical guidance for image alignment. In:
  International Workshop on Biomedical Image Registration. pp. 81--90. Springer
  (2020)

\bibitem{wang2016dueling}
Wang, Z., Schaul, T., Hessel, M., Hasselt, H., Lanctot, M., Freitas, N.:
  Dueling network architectures for deep reinforcement learning. In:
  International conference on machine learning. pp. 1995--2003 (2016)

\bibitem{watkins1992q}
Watkins, C.J., Dayan, P.: Q-learning. Machine learning  \textbf{8}(3-4),
  279--292 (1992)

\bibitem{yu2019reinforcement}
Yu, C., Liu, J., Nemati, S.: Reinforcement learning in healthcare: A survey.
  arXiv preprint arXiv:1908.08796  (2019)

\bibitem{zhong2019attention}
Zhong, Z., Li, J., Zhang, Z., Jiao, Z., Gao, X.: An attention-guided deep
  regression model for landmark detection in cephalograms. In: International
  Conference on Medical Image Computing and Computer-Assisted Intervention. pp.
  540--548. Springer (2019)

\end{thebibliography}

\end{document}